# Customer 360º Insights in Predicting Chronic Diabetes

Asish Satpathy[a*], Satyajit Behari[b]

[a] W. P. Carey School of Business, Arizona State University, Tempe, AZ, USA

[b] Univar Solutions Inc., Downers Grove, IL, USA

*Corresponding Author. Email: asish.satpathy@asu.edu

Abstract:

Chronic diseases such as diabetes are quite prevalent in the world and are responsible for a significant number of deaths per year. In addition, treatments for such chronic diseases account for a high healthcare cost. However, research has shown that diabetes can be proactively managed and prevented while lowering these healthcare costs. We have mined a sample of ten million customers' 360º data representing the state of Texas, U.S.A., with attributes current as of late 2018. The sample, received from a market research data vendor, has over 1000 customer attributes consisting of demography, lifestyle, and in some cases self-reported chronic conditions. In this study, we have developed a classification model to predict chronic diabetes with an accuracy of 80%. We demonstrate a use case where a large volume of 360° customer data can be useful to predict and hence proactively prevent chronic diseases such as diabetes.

**Introduction:**

According to the 2020 National Diabetes Statistics Report (Center of Desease Control, 2020), an estimated 34.2 million people–10.5% of the U.S. population–had diabetes in 2018. This corresponds to an increase of 1.5 million new cases of diagnosed diabetes since 2017. This diabetes epidemic is now a global phenomenon and has been reported increasingly among people in lower socioeconomic groups. Approximately 463 million adults (between age 20 and 79) were living with diabetes in 2019 and this number is expected rise to 700 million by 2045 (International Diabetes Federation, 2019). The associated health complications due to diabetes have been found to be very severe (American Diabetes Association, 2020), and this is worrisome as it may put a lot of stress on poor healthcare infrastructures, especially in developing countries. Some of the diet and lifestyle related factors that contribute to diabetes include: (1) obesity and fat distributions, (2) diet, (3) physical activity, (4) smoking, and (5) alcohol use (Hu, 2011). A drastic increase in the incidence of diabetes could be attributed to increased prevalence of obesity across all ages and ethnicities due to changes in lifestyle, a clinical condition termed as a "metabolic syndrome" (Faidon Magkos, 2009). Because these contributing factors develop and evolve over the full lifespan of patients, one could argue that they are quite difficult to manage via therapeutics. It is interesting that we did not find any past studies that potentially associated incidence of diabetes and patients' overall demography and lifestyle including, but not limited to, their interests in music and sports, finances, credit history, income, channel preference for shopping, age, education, homeownership, and much more. Past clinical studies to establish any association between incidence of chronic diseases and patients' cognitive factors have not been conclusive since a typical clinical study, performed in a hospital setting, would likely include a collection of patient's personal information by the researchers, thus likely introducing response bias as patients might not share such information honestly and correctly. Likewise, one would also expect a sampling bias when a small sample of patients is used for such study. In the current study we take a data mining approach to explore such potential associations between chronic diabetes and patients' lifestyle and demographic insights in which patients disclose their medical condition in a typical doctor visit questionnaire without divulging any personal details. We use customer 360º insights, prepared by marketing data aggregators for commercial use (mostly for target marketing), to discover such an association and develop a prediction model for the occurrence of chronic diabetes when we know a patient's lifestyle and demography in early life.

*Related Work:*

Chronic diseases such as diabetes are typically linked to prolonged periods of unhealthy habits involving poor choice of daily diet and lack of regular physical activity, in addition to existing hereditary factors. Lifestyle Medicine is a new discipline that has recently emerged as an alternative management of chronic diseases (Kushner, 2013). A recent multicohort study using 116,043 participants has found a significant association between healthy lifestyle and an increased number of disease-free-years of life (Solja T. Nyberg, 2020). Several lines of evidence support that physically active societies have a lower incidence of diabetes than less active societies, and cross-sectional studies have demonstrated inverse associations between the prevalence of type 2 diabetes and physical activity (Myers J, 2003). As a result of this inverse association, a study shows that one could expect a higher mortality rate among men for a certain group of the population with less physical activities and poor lifestyle choices (J. Larry Durstine B. G., 2013). It has been reported that people who maintain healthy and active lifestyles belong to a certain demographic and lifestyle group (Richard L. Divine, 2005), and this knowledge can be exploited to discover possible associations of certain demographics and lifestyle groups with the incidence of diabetes. However, none of these studies used large scale customer 360º insights including demography and lifestyle data to show if a prediction model could be developed to proactively prevent diabetes at an early stage.

Design of such a study with a dedicated sample is possible with a limited number of patients, however this could potentially be prone to intrinsic sampling bias in the prediction model. To improve the confidence in the prediction, one could potentially use a large amount (of the order of millions) of customer demography and lifestyle data. While this may sound very promising, it could be resource intensive and costly to create a 360° profile of a customer strictly for the purpose of these kinds of studies. In this paper, we utilize the available data from an anonymous data vendor who aggregates large scales of customer data, with more than 1000 attributes in the tens of millions of customers, from various sources.

The use of demography and lifestyle data in customer market research is quite common to understand customer purchase behavior (Krishnan, 2011). Lifestyle is an important concept used in segmenting markets and understanding target customers, which is not provided by the study of demography alone (Cahill, 2006). Studies have found that identifying the lifestyle of customers leads to development of efficient strategies for target marketing (Blumberg, 1981). Specifically, in direct marketing, where target customers are presented with a company's product and service information, detailed information regarding customers' lifestyle has proven valuable in gauging the likelihood of a response to the offer. Thanks to the proliferation of the Internet, lifestyle data is now accessible in large scale and actionable quantities and, consequently, is found to be very insightful in understanding customers when complemented with their standard demographic profile.

Traditionally, demography and lifestyle data are aggregated in a largescale manner via integrating multiple data sources by market research companies (Customer Think Corporation, 2011; Epsilon Data Management, LLC., 2021; Credit Bureau Data, Inc., 2015) to help verify or identify the market's unmet needs, identify competition, improve a company's offerings, and create more value for the organization and its customers. It should be noted that developing values from these aggregated data completely depend on the organization's needs. Data vendors aggregate these data from multiple sources and deidentify customers' personally identifiable information to protect privacy. In our study, we use one such aggregated data source with customers' demography and lifestyle information to develop a supervised classification model (Sen P.C., 2020) that predicts the incidence of diabetes. The disease information in the data has been self-reported by the customers in the database.

*Research Questions:*

Our goal for this study is to answer to the following research questions:

(1) Can we build a model to predict a customer's diabetic condition with sufficient accuracy based upon demography and lifestyle information?
(2) Can we rely on available customer data to predict chronic diseases?
(3) What will be the implication of our study for proactive management of chronic diseases such as diabetes?

While the first question can be addressed by exploring an accurate classification model that predicts diabetic condition of customers, the second question is more about validation of the developed classification model with a comparable model accuracy when scored with a large unseen 360° view of customer data. The third question will bring us insight into what role data aggregators will play in helping healthcare industries proactively prevent chronic diseases such as diabetes.

**Methodology:**

In this section we describe the data analysis methods utilized in the research. We used a standard data mining method (B. N. Lakshmi, 2011) to develop a predictive model and discover associations of customers' lifestyle with possible incidence of diabetes. The source data (with ~10 million unique records and close to 1000 unique attributes) was analyzed, cleaned, and prepared for model building as shown in the data flow diagram in Figure 1. The role of each element in the diagram is briefly discussed next.

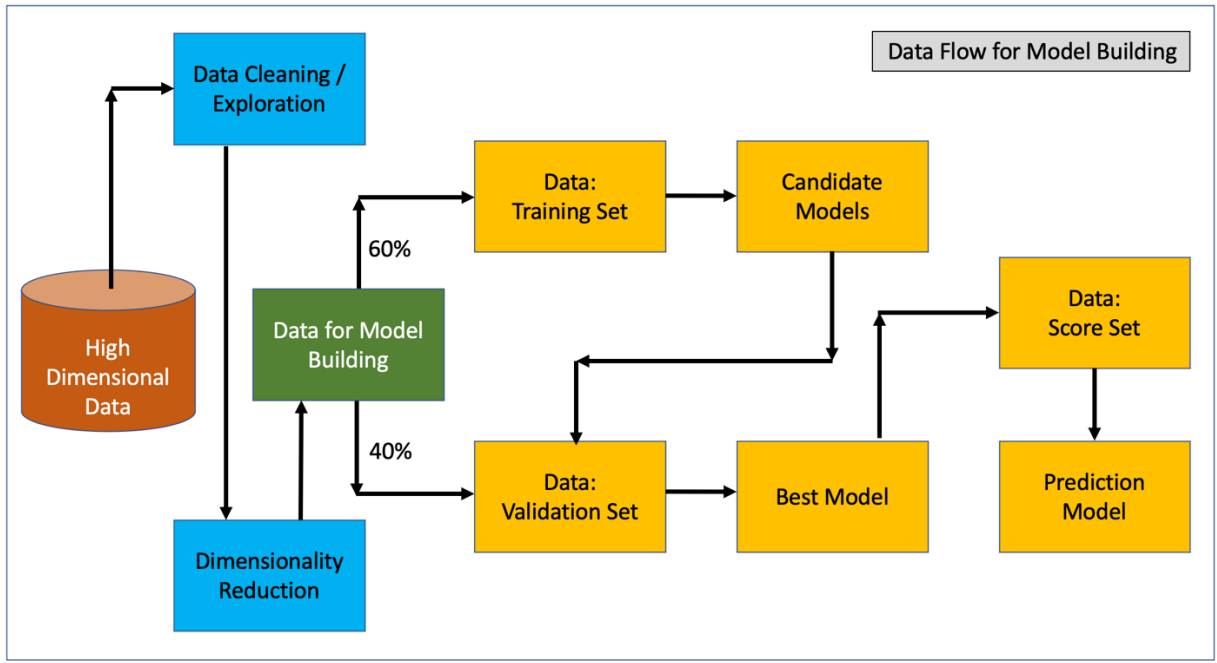

**Figure 1: Data processing flow for developing the prediction model for the incidence of diabetes**

(1) Data:

A 360° view of customers (Atkinson, 2017), built by integrating multiple data sources, is used by customer market research companies as a foundation for customer segmentation, making use of hundreds of diverse attributes. This enables organizations to connect their customer base through effective sales and promotion strategies. This results in better customer acquisition and cross-sale, in turn driving up sales and profits. The integrated data source allows marketers to create segments based on seemingly unrelated attributes and develop unique insights. The data we use in this study is an integration of several different data sources, such as credit bureaus, customers' demography and lifestyle data, to name a few. A description of some of the data variables of the sample can be found in this document (Claritas, 2019). Originally, a subset of data containing 10 million unique customers, each with over 1000 different attributes, was graciously made available to us by a reputable marketing research company. Under the terms of our nondisclosure agreement, we are not able to reveal all the details of the database schema, nor all the contents of the data. This sample is a snapshot of the 360° view of anonymous data collected from the customers in several states of the U.S.A. during the year 2018.

(2) Data Cleaning & Exploration:

Over one million records of data from the state of Texas are made available in the sample with several key categories that uniquely identified patients for our study. BASE SAS® procedures were used to clean up, impute, and visualize the data. Table 1 shows a preview of some important customer data attributes used in the analysis where customers have or have not reported on their existing chronic illnesses. Each lifestyle variable, which together form a dominant fraction among the data attributes, has a total of 99 levels, 1 being most likely and 99 being least likely an individual would belong to that category. From the pool of one million samples, we selected a sub-sample of 30,000 customer records with self-reported diabetes (signal). We also collected a mutually exclusive data sub-sample, which is derived from the same pool, of about 50,000 customers with other minor illnesses such as anxiety, allergy, osteoporosis, hearing loss, etc., who have not reported any signs of diabetes. We treated this sample as our background for the purpose of model-building. Both the signal and background samples were obtained by simple random sampling from their respective pools.

| **Demography** | **Census Data** | **Interests** | **Lifestyle** |
|---|---|---|---|
| Gender | Population | Music | Travel |
| Age | Household size | Sports | Online Preference |
| Education | Per Capita Income | Hobbies | Deposit Customer |

| **Property** | **Finance** | **Behavior** | **Credit** |
|---|---|---|---|
| Dwelling Type | Income | Channel Preference | Credit Score |
| Square Footage | Loans | Number of Purchases | Open Trades |
| Year Built | Credit Card Type | Social Media Users | Risk Predictor |

**Table 1: Sample data attributes in customers who may or may not have diabetes**

Upon exploration, we found several demography and lifestyle related predictor variables in the data sample that showed significant differences between the diabetic and non-diabetic populations. To visualize such differences, we computed proportions, a ratio of the percentage of diabetic and non-diabetic customers, for each level of predictor variables. When this proportion value is 100 for a certain category of lifestyle and demographic variable, the percentages of diabetic and non-diabetic customers showing such behavior are equal. As an illustration, Figure 2 demonstrates how a predictor, such as the *financial health* of the customer, could be used to classify between diabetic and non-diabetic patients. In this example, the proportion values are significantly different from 100 for customers with a financial health condition *below average* or *poor* and hence meet our requirements to be included in the modeling process. Figure 3 demonstrates how the variable *Length of Residence* at a given home can be used as a good predictor to classify between customers with and without diabetes. In this case, the proportion curve shows individuals with 11 or more years of residency are likely to be diabetic while those with less than 10 are not.

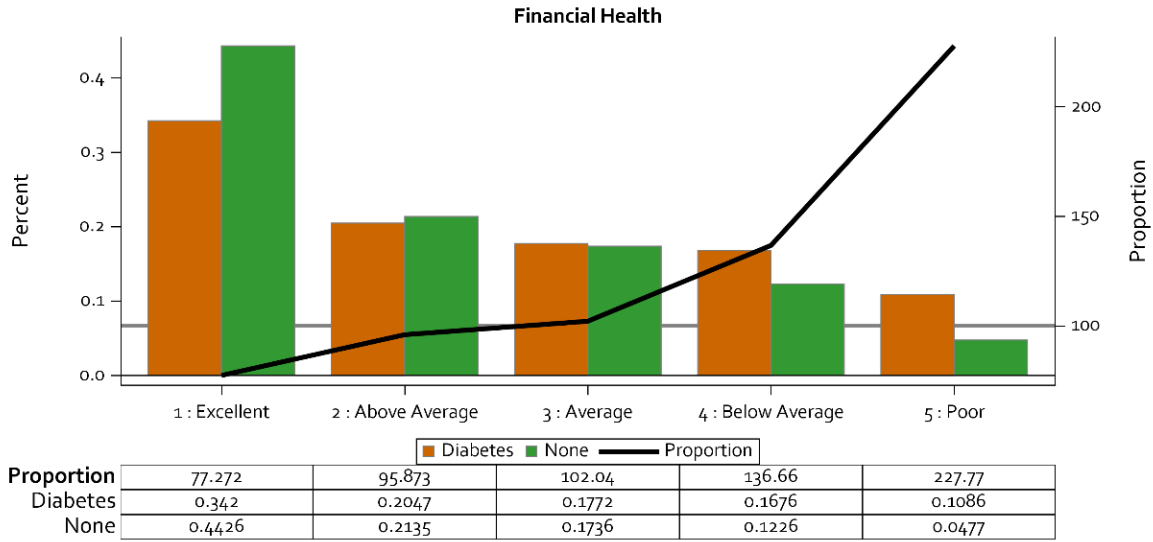

**Figure 2: Comparison of financial health of diabetic and non-diabetic customers. The proportion curve shows individuals with below average and poor financial health are disproportionately more diabetic.**

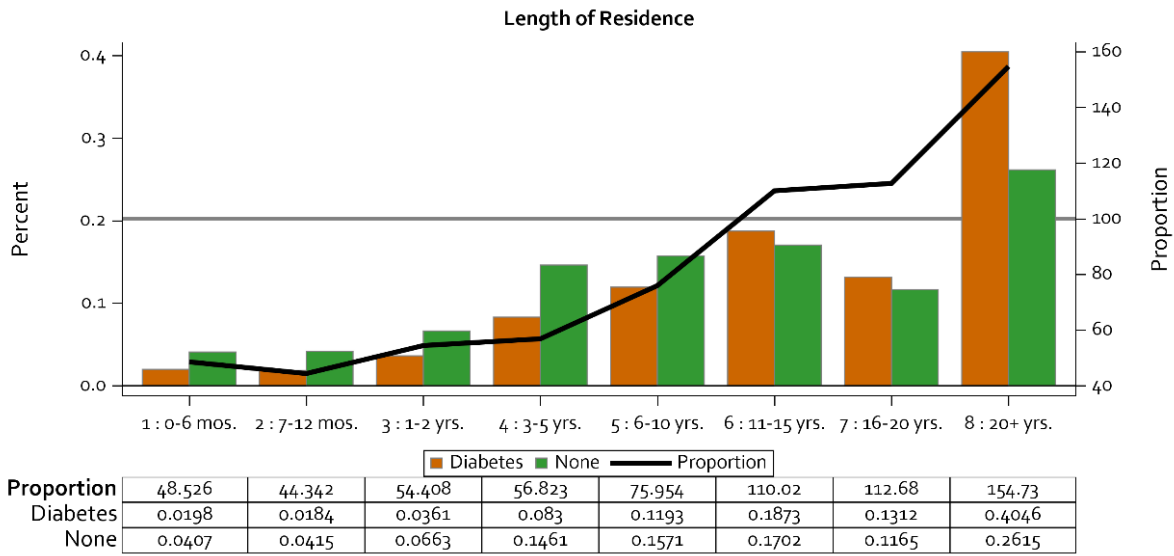

**Figure 3: Comparison of length of residence for diabetic and non-diabetic customers. The proportion curve shows individuals with 11 years or more *length of residence* in a given home are disproportionately more diabetic than those with less than 10 years.**

*(3) Data Dimensionality Reduction:*

The high dimensionality (~1000) of the data makes the analysis very challenging for data mining purposes. With the goal of reducing the dimensionality of the data to a manageable size, we applied several standard techniques to the data to reject statistically redundant/insignificant features. A summary of the selection criteria is outlined in Table 2.

In the first step, categorical variables with character levels were converted into numeric levels. Then we performed a $\chi^2$ test of binary variables with the target variable and excluded the predictors with low $\chi^2$ values. This selection resulted in pruning the dimensionality of the data to 500. Significance of the correlations between multi-valued variables with the target variable was examined using a standard t-test and the top 300 predictor variables, based on the absolute value of t-statistics, were chosen for further dimensionality reduction. The Information Value (E. Zdravevski, 2011) of categorical variables were evaluated to reduce the data dimensionality further. We selected 150 categorical predictors with Information Value between 0.03 and 0.5, ranked by their importance, in predicting the target variable. Then we performed variable clustering to find groups of variables which were most correlated within each group and least correlated with the other groups. We selected the top variable from each group based on *min (1-R² ratio)* (Aggarwal & Kosian, 2011) to be included as a predictor in the model. After this selection, we eliminated binary variables with less than 10% occupancy in the data. At this point the missing numerical values were imputed by the median of the predictor variables. At the end of this process, we performed data level clustering of categorical variables and merged levels which satisfied a cut-off value of the $\chi^2$ statistic. Finally, we were left with 50 predictor variables for each customer record in the data that we used in building the prediction model.

| Input variable selection method for target value (Diabetes: Yes or No) | Retained Dimension of Predictors |
|---|---|
| $\chi^2$ Test (based on correlation of binary variables with the target) | 500 predictors |
| t-Test (based on correlation of multivalued variables with the target) | 300 predictors |
| Select categorical variables with Information Value (IV) between 0.03 and 0.5, ranking the variables in terms of their predictive power | 150 predictors |
| (1) Perform variable clustering and select best variables based on *min* (1-R² ratio) from each cluster<br>(2) Drop binary variables with less than 10% occupancy<br>(3) Perform level clustering of categorical variables and merge levels which satisfies the $\chi^2$ cut-off with respect to the target | 50 predictors |

**Table 2: Initial variable selection method used in the analysis to reduce dimensionality of predictors in the model**

*(4) Data Split:*

We split the entire data sample (~80,000 customer records) into two mutually exclusive sets. The first set (60% of this data sample) was used to train the model and the rest (40% of the sample) was used to test our model, respectively known as the training and the validation (or holdout) set. An independent sample of 80,000 customers was selected from the state of Missouri for out-of-sample testing purpose. The use of independent data to test the model would demonstrate the generalizability of the model and its overall performance. The sample has a data dimension equal to 50 and the target value in the model has two possible outcomes: the person is (1) diabetic or (0) not diabetic.

**Model Building - Classification Analysis**

We built a classification model using SAS® Enterprise Miner with the intent to demonstrate maximum discrimination between customers with and without diabetes. We performed a stepwise logistic regression (Moore, 2005) using the Schwartz Bayesian Criterion (SAS, 2003) to identify the top significant predictors by keeping the predictor variables with p-value < 0.01, assuming the errors to be normally distributed. The continuous variables were standardized so that the Wald $\chi^2$ is comparable among all predictors. Categorical variables, with a fraction of levels appearing significant in the fit, are dummy coded to retain the significant levels. As a result, we have fewer recorded variables than all the levels of the original variables. To remove multicollinearity among the predictors in the model, we used the Lift and Cumulative Captured Response statistics to reject one of the predictors in any pair with more than 40% correlation. The same approach was repeated iteratively with the least significant variables until 16 predictor variables remained in the fit. The likelihood ratio of the *test for global null hypothesis* of this model was found to be 29337.45 leading to a p-value < 0.001. Starting with over 50 predictor variables, we eventually were left with the top 16 predictors for the model to maximize its predictability while keeping overfitting and multi-collinearity effects to the minimum.

**Model Performance:**

Model trained coefficients were used to score the target population for diabetic customers. The records are sorted by their predicted scores in descending order and divided into ten decile groups, having 10% of total customers in each decile. We then compared the Cumulative Captured Response and Lift distributions (A. Shen, 2007) for training and test samples to assess the model performance. At a given decile, "lift" is defined as the ratio of predicted rate over average rate of incidence of diabetes in the sample. As shown in Figure 4, the highest decile has a lift value of 2.4. Hence, in the first decile, compared to the 10% average identification, our model is able to identify 24% of the customers who are likely to be diabetic. This rate reduces for each successive decile.

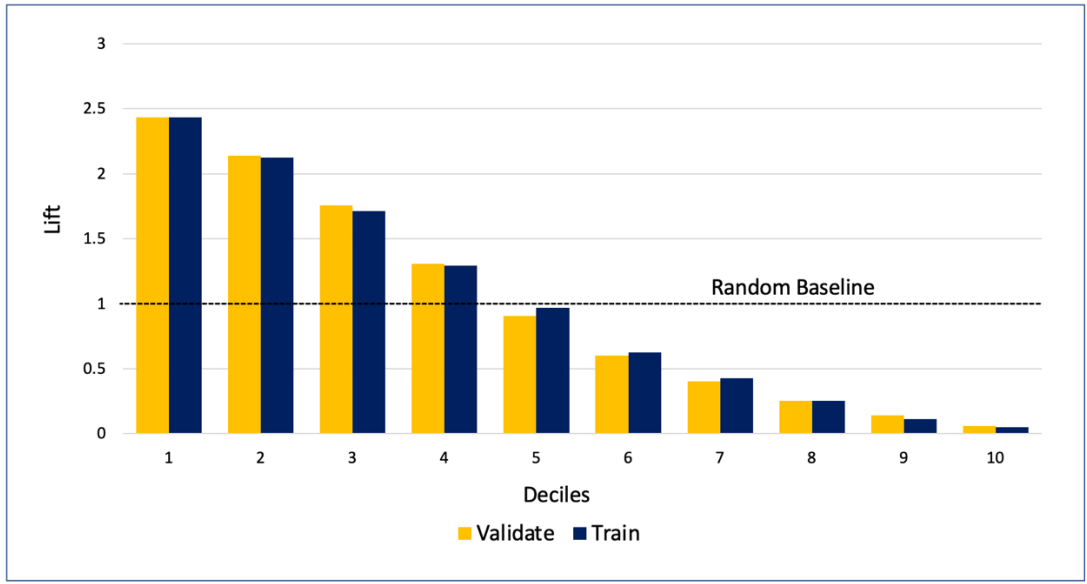

**Figure 4: Observed Lift for the model in the first decile is 2.4. The top 10% individuals based on the predictive model show 24% diabetes incidence. Both training and validation data show very similar results in the first five deciles.**

Under the assumption of approximately 35 million people per decile (that would make the total US population to be roughly 350 million), the model would correctly predict approximately 5 million (= 0.14 times 35 million) more people to have incidences of diabetes compared to a random baseline model in the first decile. Clearly our model does well to enhance events in the top four deciles, which are of most importance for targeting those customers and advising them to proactively change their lifestyles.

The Cumulative Captured Response is the rate of predicted diabetes among the customers in a cumulative decile. For example, as shown in Figure 5, the cumulative diabetes prediction rate for customers in the top five deciles is 85%. In other words, the top five deciles (approximately 175 million people of the U.S.) of the model would correctly predict 85% of diabetic patients in the U.S. The 45º line on the graph represents a random model (50-50 chance of getting a correct response as in a coin toss). This further demonstrates the reliability of our model to predict diabetic incidence for a population, given its demography and lifestyle information.

For model comparison, we have trained the same data with predictive models based on a boosted decision tree and a Neural Network. The former was found to perform marginally worse than the logistic regression model, while the latter was found to be either similar or slightly better across different deciles. For this study, we decided to adopt the logistic regression model for its simpler interpretation of relationships of the predictor variables to the target.

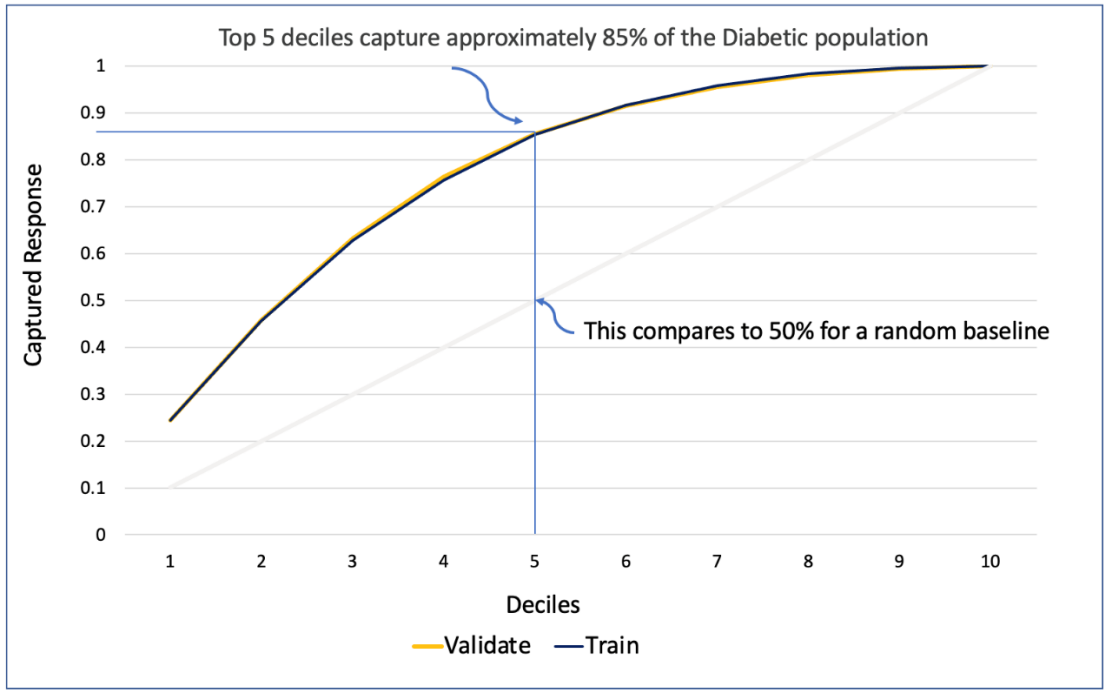

**Figure 5: Cumulative gain chart shows that top 5 deciles capture about 85% of the diabetic population in the sample. The gain chart for both Training and Validation data shows very similar results.**

**Variables in Predicting Diabetes Incidence:**

We have identified 16 significant predictor variables, as shown in Table 3, that would predict diabetes for a person given his/her demographic and lifestyle information (parameters) as indicated. The Maximum Likelihood Estimate table shows that all 16 explanatory parameters are significant in the model (p-value<0.001) at a significance level of 0.05. The logistic model formula computes the probability of incidence of diabetes, *y* (*y* = 0 if the customer does not suffer from diabetes; otherwise, *y* = 1) as a function of the values of the predictive factors shown in Table 3. If an individual suffers from diabetes, the conditional probability is given by *P(y=1 | x)* = *P(x)*, and the logistic model formula takes the form: $\log[P(x)/1-P(x)] = \beta_0 + \beta_1 x_1 + \beta_2 x_2 + \cdots + \beta_{16} x_{16}$, where *x* = (*x*$_1$, *x*$_2$, ..., *x*$_{16}$) represents the vector of 16 lifestyle related significant predictors.

Table 4 summarizes our interpretation on those predictor variables ranked from most significant to least significant. For example, according to the model, diet conscious customers have the strongest association with diabetes, which could be a result of post diagnostic behavior. The model seems to have a common thread in the predictor variables: "stress". However, it is important to remember that stress can originate from multiple sources that are not necessarily correlated with one other. Also note that these are our initial interpretations and should be verified by healthcare experts.

| Parameters | DF | Estimate | Std. Error | Wald χ² | p-value | Std. estimate | Exp (Est) |
|---|---|---|---|---|---|---|---|
| Intercept | 1 | -0.8057 | 0.0321 | 628.44 | <.0001 | | 0.447 |
| Revolving Credit Card Users | 1 | -0.3255 | 0.0204 | 253.37 | <.0001 | -0.1794 | 0.722 |
| Arts Events Patrons | 1 | 0.4395 | 0.0185 | 567.09 | <.0001 | 0.242 | 1.552 |
| Diet Conscious Customers | 1 | -1.1382 | 0.0199 | 3260.12 | <.0001 | -0.63 | 0.32 |
| Entertainment Readers | 1 | 0.5566 | 0.0179 | 965.35 | <.0001 | 0.3062 | 1.745 |
| Financial Institution Customers | 1 | 0.4008 | 0.0157 | 651.88 | <.0001 | 0.221 | 1.493 |
| Gospel Music Lovers | 1 | -0.2721 | 0.0173 | 245.97 | <.0001 | -0.1499 | 0.762 |
| Interest Checking Account Holders | 1 | 0.4253 | 0.0161 | 694.14 | <.0001 | 0.2346 | 1.53 |
| Mortgage Refinancers | 1 | -0.3379 | 0.0149 | 514.45 | <.0001 | -0.1865 | 0.713 |
| Fast Food Restaurant Users | 1 | -0.6699 | 0.0189 | 1254.88 | <.0001 | -0.3691 | 0.512 |
| Rewards Card Users | 1 | 0.7709 | 0.0187 | 1691.54 | <.0001 | 0.4249 | 2.162 |
| Senior Caregivers | 1 | -0.763 | 0.0156 | 2387.89 | <.0001 | -0.4215 | 0.466 |
| Bulk Item Shoppers | 1 | -0.5284 | 0.016 | 1095.9 | <.0001 | -0.2913 | 0.59 |
| Vacation Spenders | 1 | 0.4099 | 0.0209 | 384.73 | <.0001 | 0.2258 | 1.507 |
| Women Plus Size Clothing Buyers | 1 | -0.1949 | 0.0182 | 114.34 | <.0001 | -0.1075 | 0.823 |
| Net Worth Flag | 1 | -0.3493 | 0.0273 | 163.92 | <.0001 | | 0.705 |
| Home Value | 1 | 0.3243 | 0.0186 | 303.95 | <.0001 | 0.1821 | 1.383 |

**Table 3: Maximum likelihood estimates for the predictor parameters. We used Wald χ² values to rank these predictors in the order of importance for predicting the incidence of diabetes.**

| Rank of Ordered Predictor Variables in the Final Model | Relationship with Response | Possible Explanation |
|---|---|---|
| Diet conscious customers | Positive | More likely to become diabetic because of family history or simply showing post diagnostic behavior |
| Senior caregivers | Positive | More likely; high stress lifestyle |
| Rewards card users | Negative | Less likely; stress free lifestyle |
| Fast food restaurant users | Positive | More likely; unhealthy food habits |
| Bulk item shoppers | Positive | More likely; large volume customer |
| Entertainment readers | Negative | Less likely; stress free lifestyle |
| Financial institution customers | Negative | Less likely; affluent |
| Interest checking account users | Negative | Less likely; influenced by multiple factors |
| Arts events patrons | Negative | Less likely; affluent hobby |
| Mortgage refinances | Positive | More likely; demanding lifestyle |
| Vacation spenders | Negative | Less likely; stress free life |
| Revolving credit card users | Positive | More likely from low-income group |
| Net Worth flag | Negative | Less likely to impact people with economic stability |
| Gospel music lovers | Positive | More likely; influenced by multiple factors |
| Home value | Positive | More likely; influenced by multiple factors |
| Women plus size clothing buyers | Positive | More likely; weight factor |

**Table 4: - The final dependent variables, ranked in terms of their level of significance in the prediction model.**

**Model Evaluation:**

To test the performance of the classification model, we scored an out-of-sample data set from the state of Missouri where the true values (incidence of diabetes) are known. This data set is comprised of the same proportion of signal and background records as the training sample. The records are sorted by their predicted scores in descending order and divided into ten decile groups, containing 10% of total customers in each. Table 5 summarizes the resulting Confusion Matrix (Zeng, 2020) that we will use to determine the model accuracy. Specificity is the ability of the model to correctly identify the individuals without diabetes (true negative rate). Sensitivity is the ability of the model to correctly identify the individuals with diabetes (true positive rate). Accuracy is an indicator of the overall effectiveness of the model across the entire dataset. In the testing dataset, the logistic regression model achieved a classification accuracy of 80.02% with a sensitivity of 72.19% and a specificity of 84.76%.

- Accuracy = (TP + TN) / (TP + FP + TN + FN)

- Sensitivity = TP/ (TP + FN)

- Specificity = TN / (FP + TN)

Where TP, TN, FP and FN denote true positives, true negative, false positives and false negatives, respectively.

| | | Predicted Target | | |
|---|---|---|---|---|
| | | No Diabetes (Target = 0) | Diabetes (Target=1) | Total |
| **Decile** | **No Diabetes (Target = 0)** | | | |
| 1 | | | 761 | 761 |
| 2 | | | 2108 | 2108 |
| 3 | | | 3174 | 3174 |
| 4 | | 2892 | 1221 | 4113 |
| 5 | | 4852 | | 4852 |
| 6 | | 5326 | | 5326 |
| 7 | | 5709 | | 5709 |
| 8 | | 5926 | | 5926 |
| 9 | | 6160 | | 6160 |
| 10 | | 6275 | | 6275 |
| Total | | TN = 37140 | FP = 7264 | 44404 |
| | **Diabetes (Target=1)** | | | |
| 1 | | | 5594 | 5594 |
| 2 | | | 4247 | 4247 |
| 3 | | | 3178 | 3178 |
| 4 | | 1443 | 800 | 2243 |
| 5 | | 1503 | | 1503 |
| 6 | | 1029 | | 1029 |
| 7 | | 645 | | 645 |
| 8 | | 429 | | 429 |
| 9 | | 195 | | 195 |
| 10 | | 79 | | 79 |
| Total | | FN = 5323 | TP = 13819 | 19142 |
| | | Specificity = 84.76% | Sensitivity = 72.19% | Accuracy = 80.02% |

(Actual Target on the vertical axis)

**Table 5: Confusion Matrix for classification of customers with (Target=1) and without (Target=0) diabetes.**

**Discussion:**

The classification model developed using logistic regression techniques has merits on many fronts. We have successfully scored the model using an out-of-sample dataset consisting of customers from another state and the obtained model accuracy has clear business implications.

The predictor variables in the model indicate some interesting insights into the incidence of chronic diabetes. The top two predictor variables were further explored visually with their levels (1 being most likely, 99 being least likely the person belongs to the specific behavioral category) grouped into seven bins, as shown in

Figure 6: Charts of model variables comparing diabetic and non-diabetic customers. The varying proportion curves (in black) across different bins confirm the discriminatory power of the predictor variables to predict two different customer classes.

.

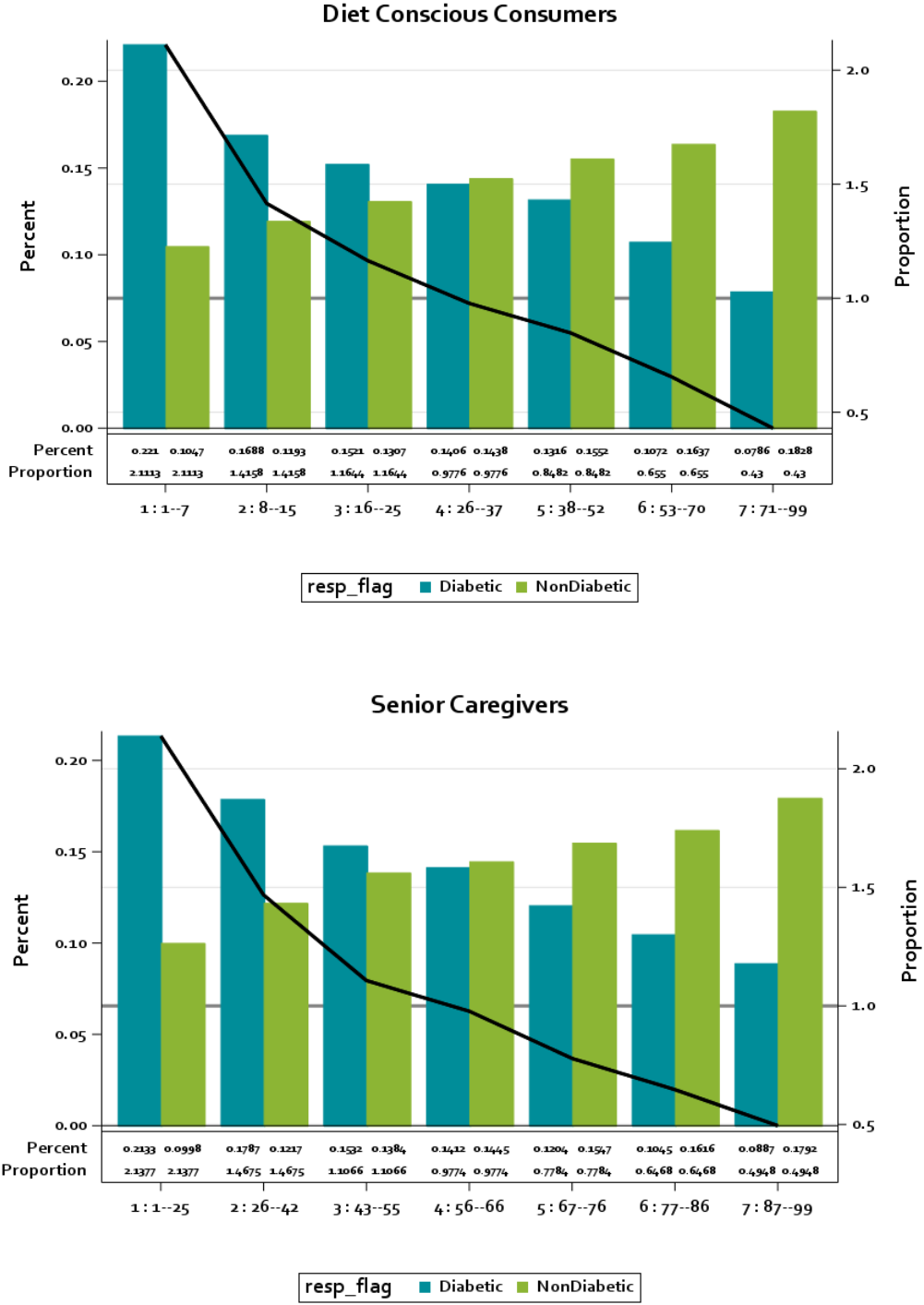

**Figure 6: Charts of model variables comparing diabetic and non-diabetic customers. The varying**

**proportion curves (in black) across different bins confirm the discriminatory power of the predictor variables to predict two different customer classes.**

Interestingly enough, almost all of the predictors in the model are related to customers' lifestyle, social behavior, and attitude towards life in general. Medical research has shown that stress, eating habits, and overall agility have significant correlations with the incidence of diabetes. In this study, we have quantified these associations to predict diabetes from available customers' 360º insights that are mostly used in marketing campaigns to acquire new customers or expand sales among existing customers.

Although our model is tested and validated to have significant predictive power for classifying customers with diabetes, we are not able to verify its accuracy on real test subjects in the absence of accompanying clinical diagnostic information. However, we believe our analysis opens a new approach to study chronic diseases that can be utilized to complement existing clinical studies. This research can be further extended to study other chronic conditions and their associations to customers' demography and lifestyle information. Since customer age is a factor in chronic diseases, further investigation is necessary to determine how the association changes given a range of customer ages. This would require additional data which we did not have access to for this study.

Currently, healthcare providers are using customers' 360º insights for marketing and operational initiatives such as (1) understanding population health and social discriminants of health, (2) marketing segmentation, (3) next best actions for patient retention, and (4) detection of fraud and abuse, to name a few. We believe that the healthcare cost can be significantly reduced by proactively targeting customers who are likely to suffer from a chronic condition and advising them to take corrective action(s). Marketing data aggregators could help healthcare providers to identify these customers and make an impact in society.

**Conclusion:**

Data mining techniques were used to analyze customers' 360º insights to predict incidence of diabetes. We have identified 16 independent demography and lifestyle related variables that could be used to predict a person's propensity to become diabetic. Most of the discovered predictors are based on customers' socio-economic status, their social behavior, and related actions associated with a certain level of stress. The implications of such results should be further studied, complemented by clinical diagnostics of those patients.

In an era of personalized medicine, our predictions could be used to guide clinical management to make decisions based on individual patients. We are motivated by the fact that most of these discovered predictors are proactively manageable to lower the incidence risk of diabetes and hope these predictors can be expanded to caution people against such adverse lifestyles. This could have an impact on people's lives and result in overall healthcare savings. However, the reader may be cautioned that initial assumptions about the data are critical to the applicability of our model.

We have demonstrated that customer market research data can be used to complement clinical research to fight chronic diseases. We are encouraged to further expand our study to predict other prevalent chronic diseases, such as hypertension and high cholesterol.